\documentclass[conference]{IEEEtran}
\IEEEoverridecommandlockouts
\usepackage{cite}
\usepackage{amsmath,amssymb,amsfonts}
\usepackage{algorithmic}
\usepackage{graphicx}
\usepackage{textcomp}
\usepackage{booktabs}
\usepackage{multirow}
\usepackage{xcolor}
\usepackage{subcaption}
\usepackage{placeins}
\usepackage{url}
\def\BibTeX{{\rm B\kern-.05em{\sc i\kern-.025em b}\kern-.08em
    T\kern-.1667em\lower.7ex\hbox{E}\kern-.125emX}}
\begin{document}
\pagestyle{empty}

\makeatletter
\newcommand{\linebreakand}{%
  \end{@IEEEauthorhalign}
  \hfill\mbox{}\par
  \mbox{}\hfill\begin{@IEEEauthorhalign}
}
\makeatother

\title{Advancing RFI-Detection in Radio Astronomy with Liquid State Machines
\thanks{This work was supported by a Westpac Future Leaders Scholarship, an Australian Government Research Training Program Fees Offset and an Australian Government Research Training Program Stipend.}
}
\author{\IEEEauthorblockN{Nicholas J. Pritchard}
\IEEEauthorblockA{\textit{International Centre for Radio Astronomy Research} \\
\textit{University of Western Australia}\\
Perth, Australia \\
0000-0003-0587-2310}
\and
\IEEEauthorblockN{Andreas Wicenec}
\IEEEauthorblockA{\textit{International Centre for Radio Astronomy Research} \\
\textit{University of Western Australia}\\
Perth, Australia \\
0000-0002-1774-5653}
\linebreakand
\IEEEauthorblockN{Mohammed Bennamoun}
\IEEEauthorblockA{\textit{School of Physics, Mathematics and Computing} \\
\textit{University of Western Australia}\\
Perth, Australia \\
0000-0002-6603-3257}
\and
\IEEEauthorblockN{Richard Dodson}
\IEEEauthorblockA{\textit{International Centre for Radio Astronomy Research} \\
\textit{University of Western Australia}\\
Perth, Australia \\
0000-0003-0392-3604}
}

\maketitle

\begin{abstract}
Radio Frequency Interference (RFI) from anthropogenic radio sources poses significant challenges to current and future radio telescopes.
Contemporary approaches to detecting RFI treat the task as a semantic segmentation problem on radio telescope spectrograms.
Typically, complex heuristic algorithms handle this task of `flagging' in combination with manual labeling (in the most difficult cases).
While recent machine-learning approaches have demonstrated high accuracy, they often fail to meet the stringent operational requirements of modern radio observatories.
Owing to their inherently time-varying nature, spiking neural networks (SNNs) are a promising alternative method to RFI-detection by utilizing the time-varying nature of the spectrographic source data. 
In this work, we apply Liquid State Machines (LSMs), a class of spiking neural networks, to RFI-detection. We employ second-order Leaky Integrate-and-Fire (LiF) neurons, marking the first use of this architecture and neuron type for RFI-detection.
We test three encoding methods and three increasingly complex readout layers, including a transformer decoder head, providing a hybrid of SNN and ANN techniques.
Our methods extend LSMs beyond conventional classification tasks to fine-grained spatio-temporal segmentation.
We train LSMs on simulated data derived from the Hydrogen Epoch of Reionization Array (HERA), a known benchmark for RFI-detection.
Our model achieves a per-pixel accuracy of 98\% and an F1-score of 0.743, demonstrating competitive performance on this highly challenging task. 
This work expands the sophistication of SNN techniques and architectures applied to RFI-detection, and highlights the effectiveness of LSMs in handling fine-grained, complex, spatio-temporal signal-processing tasks.
\end{abstract}


\section{Introduction}
Radio astronomy is a data-intensive science that heavily relies on sophisticated high-performance computing solutions to meet the demands of contemporary telescopes.
Detecting and mitigating Radio Frequency Interference (RFI) from terrestrial and satellite communications is a growing challenge.
The sheer data volume and increased sensitivity of modern and upcoming radio telescopes are pushing the limit of traditional RFI-detection methods, which often combine complex heuristic algorithms and manual inspection.

Machine learning (ML) approaches, particularly convolutional neural networks (CNNs), have been applied to RFI-detection with some success \cite{akeret_radio_2017, yang_deep_2020, vafaeisadr_deep_2020}.
However, these methods typically treat RFI-detection as a binary segmentation task on two-dimensional spectrograms and can be computationally expensive, limiting their real-time applicability in operational settings \cite{mesarcik_learning_2022}.

Spiking Neural Networks (SNNs) are time-varying systems inspired by biological neural systems \cite{trappenberg_fundamentals_2010}.
Exploiting SNN's dynamic nature to process the temporal nature of radio astronomy observations is an appealing and promising alternative to ANN-based RFI-detection approaches.
Recent studies have begun to explore the application of SNNs to RFI-detection by leveraging their dynamic nature to handle spatio-temporal data effectively \cite{pritchard_rfi_2024, pritchard_supervised_2024, pritchard_spiking_2024}. 

Liquid State Machines (LSMs), a form of reservoir computing with SNNs \cite{maass_real-time_2002}, have shown potential in various signal processing tasks, including temporal pattern recognition \cite{goodman_spatiotemporal_2006}.
Unlike other deep learning approaches, which require training all parameters in a network, LSMs require only training a readout layer, significantly reducing training complexity and computational cost.
Their capacity to handle time-varying inputs coupled with lighter-weight training requirements makes them appealing candidates for addressing the challenges of RFI-detection in radio astronomy, where radio spectrograms evolve dynamically over time.

In this work, we extend the application of LSMs to the domain of RFI-detection by:
\begin{enumerate}
    \item Applying LSMs to RFI-detection formulated as time-series segmentation with synthetic data emulating the Hydrogen Epoch of Reionization Array (HERA) telescope.
    \item Utilizing second-order Leaky Integrate-and-Fire (LiF) spiking neurons to RFI-detection with SNNs testing three encoding methods: latency, rate and direct current encoding.
    \item Testing three readout layer configurations: linear, ReLU and transformer-decoder. Utilizing the spiking behavior of reservoir as memory for the transformer layer.
\end{enumerate}

Our results demonstrate that LSMs can achieve competitive performance on this challenging segmentation task, expanding the collection of tasks to which they have been successfully applied.
This work advances the sophistication of SNN-based methods for RFI-detection in radio astronomy and establishes a precedent for applying LSMs to complex spatio-temporal problems. By leveraging LSMs' inherently temporal nature, we aim to develop more efficient data-driven and real-time RFI-detection methods.
\section{Related Work}
RFI-detection and mitigation is essential to operating contemporary radio telescope observatories.
RFI-detection, termed `flagging' requires ingesting partially processed `visibility' data from the telescope's interferometric correlator and outputting a boolean mask (flag) indicating the presence of RFI at a particular time-step in a particular channel.
Contemporary RFI-detection methods rely on cumulative sum algorithms to identify regions of increased power within spectrograms in addition to expertly tuned heuristics and filters \cite{offringa_aoflagger_2010}.
While these methods are operationally efficient, they require ongoing fine-tuning, expert oversight, and, in challenging cases, manual labeling, making them difficult to scale for next-generation radio telescopes such as the Square Kilometer Array (SKA) \cite{vermij_challenges_2015}.

The sheer volume of available historical observation data and the promise of increasing future observation data volumes provide strong motivation for investigating data-driven and machine-learning techniques to aspects of radio astronomy processing.
Many have tackled RFI-detection with machine learning as a two-dimensional segmentation problem similar to traditional algorithms.
UNet-like architectures, in particular, have been widely investigated and achieve high accuracy in segmenting spectrograms using the outputs of well-tuned traditional algorithms as training supervision \cite{akeret_radio_2017, yang_deep_2020, vafaeisadr_deep_2020}.
To address the undesirable reliance on using existing flagging techniques as supervision targets, unsupervised nearest-neighbor methods \cite{wolfaardt_machine_2016} and auto-encoder-based anomaly detection methods have additionally been investigated \cite{mesarcik_learning_2022, vanzyl_remove_2024}.
Most recently, preliminary work into deploying vision-transformer models \cite{ouyang_hierarchical_2024} shows promise, albeit at a high computational cost. Du Toit et al. \cite{dutoit_comparison_2024} provide a comprehensive comparison of deep learning approaches to RFI-detection. While many methods provide excellent detection performance, they incur high operational requirements compared to traditional methods.

The promise of marrying the time-varying nature of radio astronomy observations and SNNs has been known to be appealing for several years \cite{scott_evolving_2015, kasabov_evolving_2016}; however, few have investigated applying SNNs and neuromorphic computing to any tasks in radio astronomy observations or RFI-detection.
Pritchard et al. \cite{pritchard_rfi_2024} provided the first attempt to apply SNNs to RFI-detection by adapting an auto-encoder-based anomaly detection scheme \cite{mesarcik_learning_2022} through ANN2SNN conversion.
While accurate, this method only reduced the memory requirements compared to the original technique rather than providing improved detection performance or time efficiency.
Later, the same team moved to supervised training small SNNs by reformulating RFI-detection as a time-series segmentation problem \cite{pritchard_supervised_2024}.
We borrow the problem formulation and datasets of this later work.
While pioneering, these two works only use first-order leaky-integrate and fire (LiF) neurons and do little to consider biological plausibility or inspiration in their approaches.
We aim to partially address these shortcomings in this work.

Liquid State Machines (LSMs) offer a biologically plausible time-series processing paradigm based on SNNs \cite{maass_real-time_2002}.
Despite their promise and elegance, correctly tuning the liquid state and neuron dynamics are key to achieving a suitably dynamic but not completely chaotic liquid \cite{goodman_effectively_2005}.
LSMs have previously been shown to handle complex time-series processing tasks such as robot arm path prediction \cite{burgsteiner_movement_2007}, generalized pattern recognition \cite{goodman_spatiotemporal_2006}, uni-variate time-series classification \cite{gaurav_reservoir_2023}, and speech recognition reaching performance competitive with backpropagation through time trained SNNs \cite{deckers_extended_2022}.
Moreover, the simplicity in defining the un-trained liquid reservoir makes LSMs amenable to hardware acceleration \cite{zhang_digital_2015, saraswat_hardware-friendly_2021, wang_lsmcore_2022, lin_resistive_2023}.
RFI-detection, formulated as a time-series segmentation task, provides an exceedingly challenging task for LSMs, which have largely been focused on classification problems.
Moreover, this work is the first to explore the application of LSMs to radio astronomy and RFI-detection in particular.
By extending LSMs to perform fine-grained RFI-detection, we aim to investigate their utility in addressing the unique challenges posed by radio astronomy.
\section{Methods}
We continue from the RFI-detection formulation Pritchard et al. \cite{pritchard_supervised_2024} present, which casts what is traditionally treated as a two-dimensional image-segmentation problem as a multi-variate time-series segmentation problem more suited for SNN evaluation.
By iterating on the model architectures tested, we uncover performance in spike-encoding methods previously considered unsuitable for this task.
Our main contribution is expanding the array of architectures, and neuron models applied to this formulation of RFI-detection.
Fig. \ref{fig:summary_figure} outlines our overall approach encompassing comparing encoding methods and LSM readout layers.
The example encodings show that rate encoding exhibits the most sparsity, latency encoding provides an offset perturbation in each time step, and direct encoding retains input magnitude.
Including a second-order LiF neuron model introduces complexity and temporal dynamics, which we balance against the biologically inspired LSM architecture, which is noted for its relative training ease.
Comparing increasingly complex readout layers further expands the insight into this problem space and broadens the set of tasks to which LSMs have been applied.

\begin{figure*}[!tb]
    \centering
    \includegraphics[width=515pt, height=250pt]{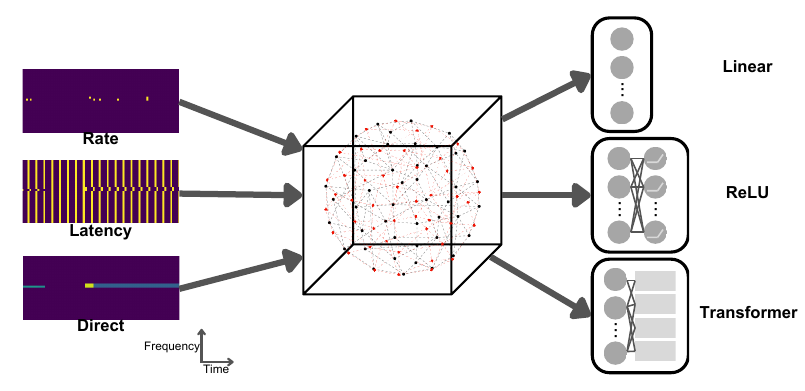}
    \caption{Overall approach to RFI-detection with liquid state machines. One of several encoding methods (left) is fed into a recurrently connected LSM (center), the output of which feeds into one of three decoding architectures (right).}
    \label{fig:summary_figure}
\end{figure*}

This section outlines the time-series segmentation formulation of RFI-detection with SNNs, the dataset tested and pre-processing techniques applied, our LSM definition, parameters, and spiking neuron model. It briefly introduces the encoding methods tested, the readout layer variations tested, the hyper-parameter-optimization routine, and finally, the compute, training, and simulation environment details.
\subsection{RFI-Detection with SNNs}
We adopt the time-series segmentation approach for RFI-detection as Pritchard et al. \cite{pritchard_supervised_2024} propose.
In this formulation, the loss function supervises the spiking network to output a boolean mask of `flags' as:
\begin{equation}\label{eq:newformulation}
    \mathcal{L}_{sup} = 
    min_{\theta_n}(
        \Sigma_t^T
        \mathcal{H}(
            m_{\theta_n}(
                E(V(\upsilon, t, b))
            ), 
            F(G(\upsilon, t, b))
        )
    )
\end{equation}
where $V$ is the original visibility data, $G$ is the supervised ground truth, $E$ and $F$ are spike-encoding and decoding functions, respectively, $\theta_n$ are the parameters of some classifier $m$ and $\mathcal{H}$ is a comparator function between the supervision masks and output of the network. Finally, $\upsilon$ is the visibility frequency, $t$ the time and $b$ baseline (pair of antennae) in the original visibility data.
This formulation introduced the concept of exposure-time $e$.
To allow the network dynamics to produce an output per time step (in the original spectrogram), each time step is expanded (and correspondingly compressed) by $E$ and $F$.

Three spike-encoding methods were tested:
\begin{itemize}
    \item \textbf{Latency encoding} encodes inputs and outputs as a spike delay proportional to its value as described previously in \cite{pritchard_supervised_2024}.
    \item \textbf{Rate encoding} converts input and output intensity into spike frequency. Gradient updates are calculated on the average activity of the network per spectrogram time-step, unlike in \cite{pritchard_supervised_2024}.
    \item \textbf{Direct encoding} is an approach where $E$ is an identity function, treating the visibility values as driving currents for each input channel, repeating the value for the desired exposure time. Output decoding is the same as rate decoding.
\end{itemize}
Previous methods making use of this formulation provide supervision on the output spike-rasters directly \cite{pritchard_supervised_2024, pritchard_spiking_2024}; in this work, we provide supervision on the decoded segmentation decisions, providing a stronger gradient influence.
We investigate several encoding methods alongside readout layer designs as they encourage the network to represent information in different ways; some vastly more effective than others.
There is no a priori way to know which method will perform best and it could be the case that a particular encoding method paired with a particular readout architecture performs especially well.
\subsection{Dataset}
The dataset we use originates from the HERA simulator \cite{deboer_hydrogen_2017} as prepared in \cite{mesarcik_learning_2022} and has been used to benchmark several other RFI-detection algorithms \cite{dutoit_comparison_2024}.
As is common in other machine-learning-based RFI-detection methods, the original $512 \times 512$ pixel spectrograms are split into $32 \times 32$ chunks to shrink the model size significantly.
The specific version this work uses employs the divisive normalization techniques introduced in \cite{pritchard_spiking_2024} as it has been shown to improve performance significantly.
Like the original, this dataset contains 420 training and 140 test spectrograms with an overall RFI contamination of around 3\%. See \cite{pritchard_supervised_2024, mesarcik_learning_2022} for a more detailed description of this dataset.
\subsection{Liquid State Machine Parameters}
The LSM architecture consists of a reservoir of recurrently connected spiking neurons with a random distribution of parameters to encourage a dynamic response to various inputs.
In line with known literature \cite{deckers_extended_2022}, we initialize the reservoir with an 80/20 ratio of excitatory and inhibitory connections.
We use second-order Leaky Integrate and Fire (LiF) neurons characterized by the dynamics:
\begin{equation}
\begin{aligned}
    I_{syn}^{t+1} = I_{syn}^{t} + S_{in}(t) + S_{rec} \cdot
    W_{rec} \\
    I_{syn}^{t+1} = I_{syn}^{t} \cdot \exp(-dt / \tau_{syn}) \\
    V_{mem}^{t+1} = V_{mem}^{t} \cdot \exp(-dt / \tau_{mem}) \\
    V_{mem}^{t+1} = V_{mem}^{t} + I_{syn}^{t} + b + \sigma \zeta(t)
\end{aligned}
\end{equation}
as per Rockpool's implementation \cite{muir_rockpool_2019} where $V_{mem}$ is the membrane voltage potential, 
$I_{syn}$ is the synaptic current,
$S_{in}$ are the incoming spikes,
$S_{rec}$ are recurrent spikes,
$W_{rec}$ is the current weight,
$\tau_{syn}$ is the synaptic time-constant,
$\tau_{mem}$ is the membrane time-constant,
$b$ is a bias current,
$\zeta(t)$ is a Wigner random noise function, and the neuron membrane is reset by subtraction.
We initialize the reservoir neurons with time constants $\tau$ ranging uniformly from $0.001$ to $0.01$ and uniformly random recurrent weights.
\subsection{Readout Layer Variations}
We only train the readout layer in LSMs, leveraging the reservoir's dynamics.
We evaluated three readout architectures of increasing complexity, each mapping directly to the 32 output channels:
\begin{itemize}
    \item \textbf{Linear Layer}: A single layer of weights.
    \item \textbf{Linear + ReLU Layer}: Linear weights with ReLU-activated artificial neurons.
    \item \textbf{Transformer Decoder}: A transformer decoder block with four attention heads, ReLU-activated neurons and using the spiking history as memory.
\end{itemize}
Table \ref{tab:networkarch} summarizes our model architectures, in all cases the input and output is 32 spectrogram channels.
Since we train only the readout layer, these models significantly reduce computational overhead compared to other similarly scaled SNNs or ANNs, making them attractive for real-time applications.
\begin{table}[!tb]
\centering
\caption{Neural Network Architectures. $N_{LSM}$ refers to the size of the reservoir.}
\label{tab:networkarch}
\begin{tabular}{@{}lll@{}}
\toprule
\textbf{Layer} & \textbf{Input} & \textbf{Output} \\ \midrule
Input        & 32       & $N_{LSM}$             \\
Reservoir        & $N_{LSM}$            & $N_{LSM}$     \\
\begin{tabular}[l]{@{}l@{}}Readout (Linear,\\ ReLU, Transformer)\end{tabular} & $N_{LSM}$ & 32 \\
\bottomrule
\end{tabular}
\end{table}
\subsection{Training Environment}
All experiments were conducted using SynSense's Rockpool library \cite{muir_rockpool_2019} which is based on PyTorch, in combination with PyTorch Lightning \cite{falcon_pytorch_2019} for efficient data-driven parallelism.
Code is publicly available online\footnote{\url{https://github.com/pritchardn/SNN_RFI_LSM}}.
All trials were optimized with Adam with an initial learning rate of 0.0001 scheduled to reduce on validation loss plateaus by a factor of 0.5 and patience of 10.
All models were trained for 100 epochs except hyper-parameter trials, which were trained for 50 epochs and 10\% of the original training set.
These training parameters are in-line with previous investigations \cite{pritchard_supervised_2024, pritchard_spiking_2024} using the same datasets, although it may be possible to extract further performance with more epochs.
\subsection{Hyper-Parameter Optimization}
We use Optuna's \cite{akiba_optuna_2019} tree-structured Parzan estimation algorithm to optimize for F1-Score.
We select to optimize for F1-Score as it reflects the inherent imbalance present in the RFI-detection task; RFI typically comprises at most 10\% of all pixels, and therefore, while per-pixel accuracy may be high, this can be boosted by a large proportional of true-negative predictions.
Table \ref{tab:hyperparams} describes the parameters selected for optimization and their ranges.
Table \ref{tab:hyperparams_results} contains the final parameters selected for each spike-encoding method and readout layer over 25 trials each.
Each of the 25 trials is run on 10\% of the original training set and trained for 50 epochs each.
\begin{table}[!tb]
    \centering
    \caption{Ranges for parameters included in hyper-parameter optimization routines.}
    \label{tab:hyperparams}
    \begin{tabular}{@{}ll@{}}
    \toprule
    Parameter  & Range \\ \midrule
    Input Sparsity & [0.0, 1.0]        \\
    Exposure       & [1, 2, 4, 8, 16, 32]         \\
    Reservoir Size & [512, 1024, 2048, 4096, 8192]      \\ \bottomrule
    \end{tabular}
\end{table}
\begin{table}[!tb]
\caption{Hyper-parameter optimization results}
\label{tab:hyperparams_results}
\begin{tabular}{@{}cccccc@{}}
\toprule
Readout Layer                & \begin{tabular}[c]{@{}c@{}}Encoding\\ Method\end{tabular} & F1-Score  & \begin{tabular}[c]{@{}c@{}}Input\\ Sparsity\end{tabular} & Exposure  & \begin{tabular}[c]{@{}c@{}}Reservoir\\ Size\end{tabular} \\ \midrule
\multirow{3}{*}{Linear}      & Rate                                                      & 0.55      & 0.19                                                     & 16        & 8192                                                     \\
                             & Latency                                                   & 0.1       & 0.298                                                    & 4         & 4096                                                     \\
                             & Direct                                                    & 0.654     & 0.074                                                    & 1         & 4096                                                     \\ \midrule
\multirow{3}{*}{ReLU}        & Rate                                                      &           0.552&                                                          0.178&           16&                                                          1024\\
                             & Latency                                                   & 0.100           & 0.377                                                         &  16         & 4096                                                         \\
                             & Direct                                                    & 0.481& 0.149& 4& 2048\\ \midrule
\multirow{3}{*}{Transformer} & Rate                                                      & 0.778     & 0.285                                                    & 4         & 2048                                                     \\
                             & Latency                                                   &           0.094&                                                          0.382&           2&                                                          4096\\
                             & Direct                                                    &           0.785&                                                          0.230&           32&                                                          1024\\ \bottomrule
\end{tabular}
\end{table}

Optuna's fANOVA importance evaluator found input sparsity to be the most important variable by a significant margin, followed by exposure time and liquid size in all trials.
Commenting briefly on the trial results, latency encoding exhibits the weakest performance, followed by rate encoding and then direct encoding.
Input sparsity, in line with literature expectations \cite{deckers_extended_2022}, falls below 40\% connectivity in all cases.
Moreover, latency encoding requires the highest input density, followed by rate encoding and, finally, direct encoding, indicating a potential relationship between encoding efficiency.
As the readout layers increase in complexity, so too does input density.
Exposure and reservoir size otherwise span the entire range of inputs, reflecting their lessened importance.
Moreover, performance improves as the readout layer complexity increases resulting in direct encoding with a transformer readout layer providing the most promising parameters.
\section{Results}
Table \ref{tab:results} contains the averaged performance of each encoding method and readout layer over five trials.
\begin{table*}[!tb]
\centering
\caption{Final results averaged over 5 trials presented as mean and standard deviation. Best scores are in bold.}
\label{tab:results}
\begin{tabular}{@{}cclclclclc@{}}
\toprule
Readout Layer                & \begin{tabular}[c]{@{}c@{}}Encoding\\ Method\end{tabular} & \multicolumn{2}{c}{Accuracy} & \multicolumn{2}{c}{AUROC} & \multicolumn{2}{c}{AUPRC} & \multicolumn{2}{c}{F1-Score} \\ \midrule
\multirow{3}{*}{Linear}      & Rate                                                      & 0.9742          & 0.0019                  &  0.7152       & 0.0220                & 0.6429        & 0.0293                & 0.5569         & 0.0416                  \\
                             & Latency                                                   & 0.3599         & 0.012                  &  0.4959       & 0.0052  & 0.3501        &   0.0063    & 0.0795         &  0.0030         \\
                             & Direct                                                    & 0.9772         &  0.0049                 &  0.7226       &   0.0576  & 0.7087        & 0.0552                & 0.5976   & 0.1058                  \\ \midrule
\multirow{3}{*}{ReLU}        & Rate                                                      & 0.9748         & 0.0049                  &  0.7335       & 0.0538                & 0.6778        & 0.0546                & 0.5942         & 0.0937                  \\
                             & Latency                                                   & 0.9544         & 0.0036                  & 0.5021        & 0.0018                & 0.1317        & 0.1724                & 0.0759          & 0.0013                  \\
                             & Direct                                                    & 0.9781         & 0.0011         &        0.7085 & 0.0194       & 0.7175  & 0.0145       & 0.6658  & 0.0237         \\ \midrule
\multirow{3}{*}{Transformer} & Rate                                                      & 0.9804         & 0.0007                  & 0.8081        & 0.0068                & 0.7588        &  0.0038               &  0.7108        &  0.0088                 \\
                             & Latency                                                   & 0.2412         & 0.0880                  & 0.5079        &  0.0121        & 0.4237   & 0.0339                 & 0.0790         &  0.0032                 \\
                             & Direct                                                    & \textbf{0.9824}         & 0.0010                 & \textbf{0.8318}        & 0.0107                 & \textbf{0.7808}        & 0.0082                & \textbf{0.7432}          & 0.0144                  \\ \bottomrule
\end{tabular}
\end{table*}
We compare methods in Accuracy, Area Under the Receiver-Operator Curve AUROC, Area Under the Precision-Recall Curve AUPRC and F1-Score, paying greater attention to class balanced-performance metrics AUPRC and F1-Score.
We discuss performance differences between encoding techniques and readout layer complexity and compare them to existing SNN-based RFI-detection methods.
\subsection{Encoding Comparison}
Generally, performance differences between the encoding methods in Table \ref{tab:results} match the performance differences seen in the hyper-parameter trials in Table \ref{tab:hyperparams_results}.
Latency encoding performs significantly worse than rate or direct encoding in all cases by quite some margin.
We hypothesize this is due to the freezing of the reservoir parameters since the readout layer can only affect the precise spike timings so much.
This deviates from prior results \cite{pritchard_supervised_2024}, which found latency encoding to perform better than other methods and rate encoding in particular.
We expect this difference to arise from the fact that prior works directly supervised the output spike rasters rather than the decoded predictions, as we do in this work.
Rate and direct encoding methods perform markedly better and similarly to each other, something we would expect, given they share a common decoding technique.
Direct encoding, however, performs best independently of the readout layer used, providing an $\approx$ 5-10\% improvement in AUPRC and F1-score metrics over rate encoding.
Moreover, we see that latency encoding is subject to the greatest variation in performance with the exception of the F1-Score of direct encoding with a linear decoding layer which is likely the result of employing a short exposure time of a single time-step.
These results suggest that under an LSM framework, rate decoding with direct encoding provides the best performance in contrast to existing work \cite{pritchard_supervised_2024}.
\subsection{Readout Layer Comparison}
We additionally compare the differences between increasingly complex readout layers.
We see relative subsequent improvements in accuracy between linear, ReLU and transformer readouts in rate and direct encoding.
AUROC improves by 0.02 and then 0.07 for rate encoding across ReLU and transformer readouts, but we see a regression of 0.02 followed by an improvement of 0.11 (over the linear readout result) for direct encoding.
Moving from linear to ReLU and transformer layers yields 0.03 and 0.08 incremental improvements in the AUPRC score for rate encoding and 0.01 and 0.07 for direct encoding, respectively.
Finally, F1 Scores improve by 0.08 when using a ReLU readout, 0.11 when moving to a transformer readout for rate encoding, and incrementally 0.07 and 0.08 for direct encoding.
In combination with the increased input density suggested by the hyper-parameter optimizations in Table \ref{tab:hyperparams_results}, it is clear that the increasingly complex readout layers do provide significant performance improvements and, more notably, that the spiking output of the reservoir serves as a useful memory in the transformer decoder.
\subsection{Comparison to Related Methods}
Table \ref{tab:results_comparison} compares other ANN and SNN-based RFI-detection methods on this dataset. 
The other methods exhibit superior performance, however LSMs are typically employed for classification problems and not complex time-series segmentation tasks.
We also note that the ANN and ANN2SNN auto-encoder examples \cite{mesarcik_learning_2022, pritchard_rfi_2024} are convolutional and therefore used at least an order of magnitude more parameters.
The rate-coded LSM with a transformer layer readout shows results comparable to a latency-encoded SNN-MLP \cite{pritchard_supervised_2024}.
While the ANN2SNN auto-encoder-based method demonstrates the best F1-Score, it requires 512 exposure steps per $32 \times 32$ spectrogram patch and a significantly more intensive CNN-based auto-encoder architecture \cite{pritchard_rfi_2024}.
While our LSM approach does not surpass the best-performing ANN2SNN or SNN-MLP methods, the training process is significantly simpler as the spiking neurons do not need to be trained.
Fig. \ref{fig:example} illuminates why performance from the LSM may lag that of the current state-of-the-art; the LSM struggles to deal with instantaneous, wide-band RFI.
This may be resolved by employing a second reservoir with dynamics tuned to handle faster changing dynamics and integrating the output of both reservoir columns.
\begin{table}[!tb]
\centering
\caption{Results compared to other ANN and SNN-based RFI-detection methods}
\label{tab:results_comparison}
\begin{tabular}{@{}ccccc@{}}
\toprule
Work      & Method            & AUROC & AUPRC & F1-Score \\ \midrule
\cite{mesarcik_learning_2022}   & ANN Auto-Encoder  & 0.981 & 0.927 & 0.910 \\
\cite{pritchard_rfi_2024}      & ANN2SNN - Auto-Encoder          & 0.944 & 0.910 & 0.953    \\
\cite{pritchard_supervised_2024}     & BPTT - SNN-MLP & 0.929 & 0.785 & 0.761    \\
\cite{pritchard_spiking_2024}  & BPTT - SNN-MLP & 0.968 & 0.914 & 0.907    \\
This work & LSM               & 0.842     & 0.781     & 0.743        \\ \bottomrule
\end{tabular}
\end{table}
\begin{figure}[!tb]
    \centering
    \begin{subfigure}{\columnwidth}
        \centering
        \includegraphics[width=\textwidth, height=1.5in, keepaspectratio]{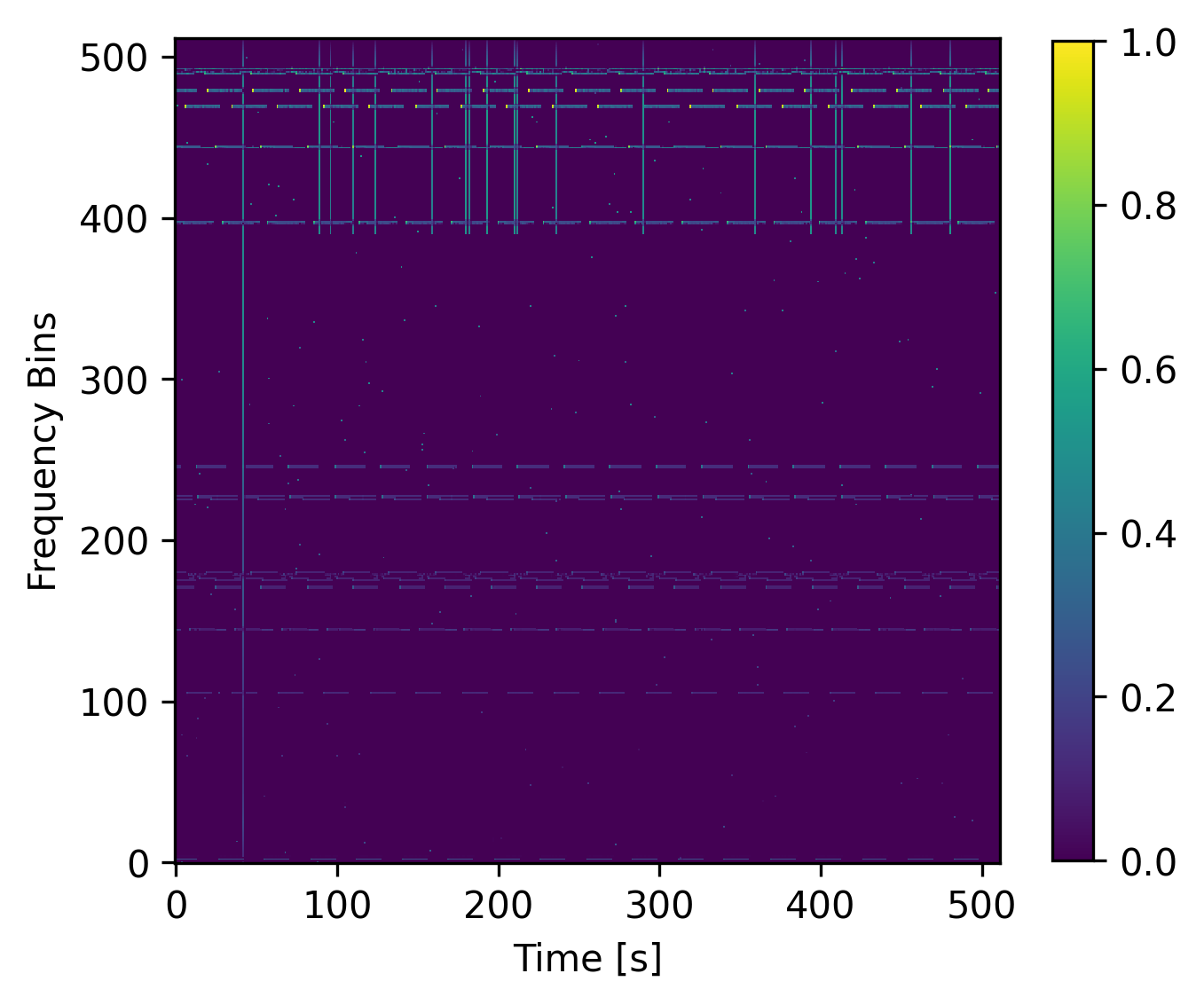}
        \caption{Original spectrogram.}
        \label{fig:res:orig}
    \end{subfigure}
    \begin{subfigure}{\columnwidth}
        \centering
        \includegraphics[width=\textwidth, height=1.5in, keepaspectratio]{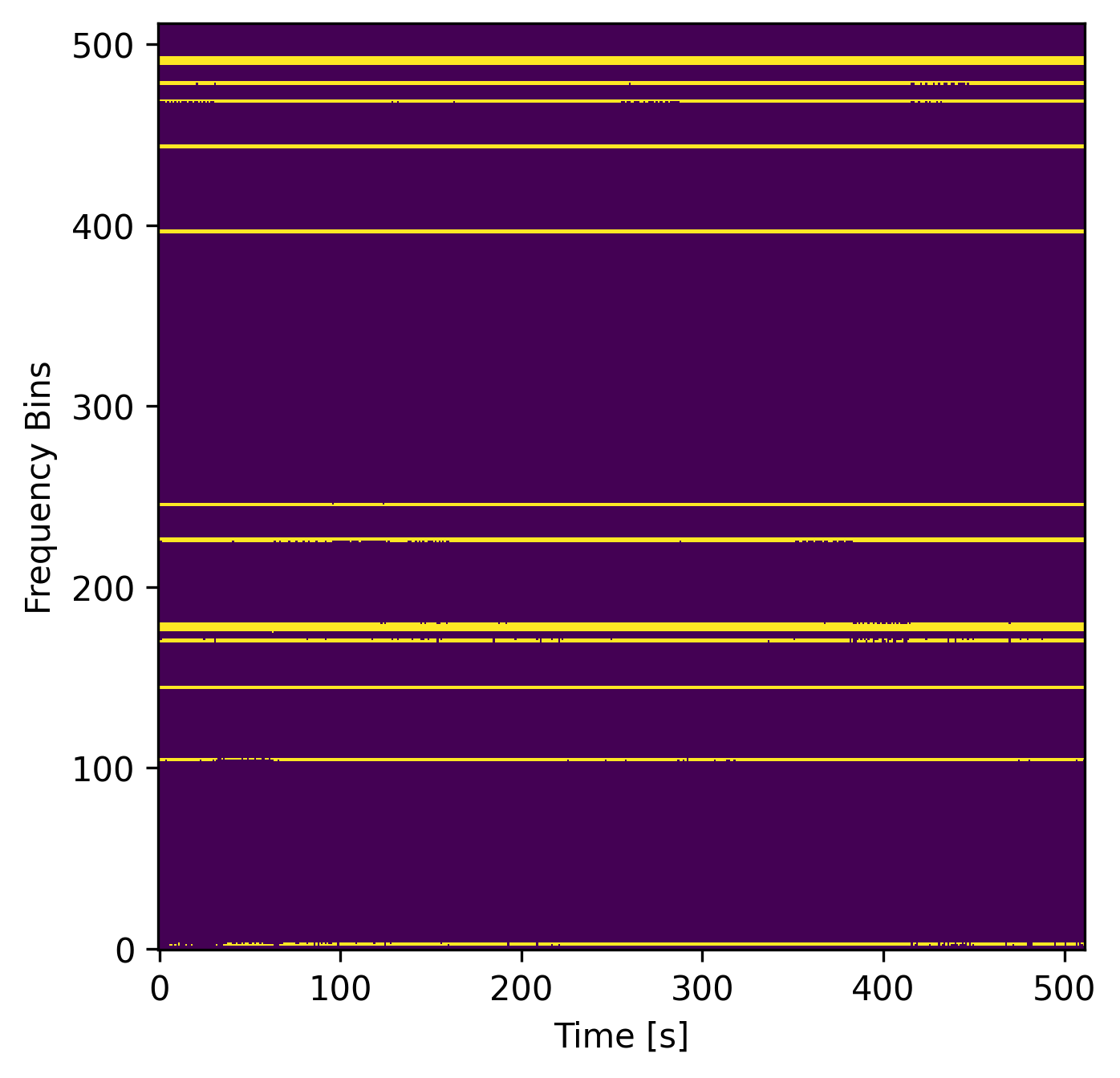}
        \caption{Latency-based inference.}
        \label{fig:res:inf}
    \end{subfigure}
    \begin{subfigure}{\columnwidth}
        \centering
        \includegraphics[width=\textwidth, height=1.5in, keepaspectratio]{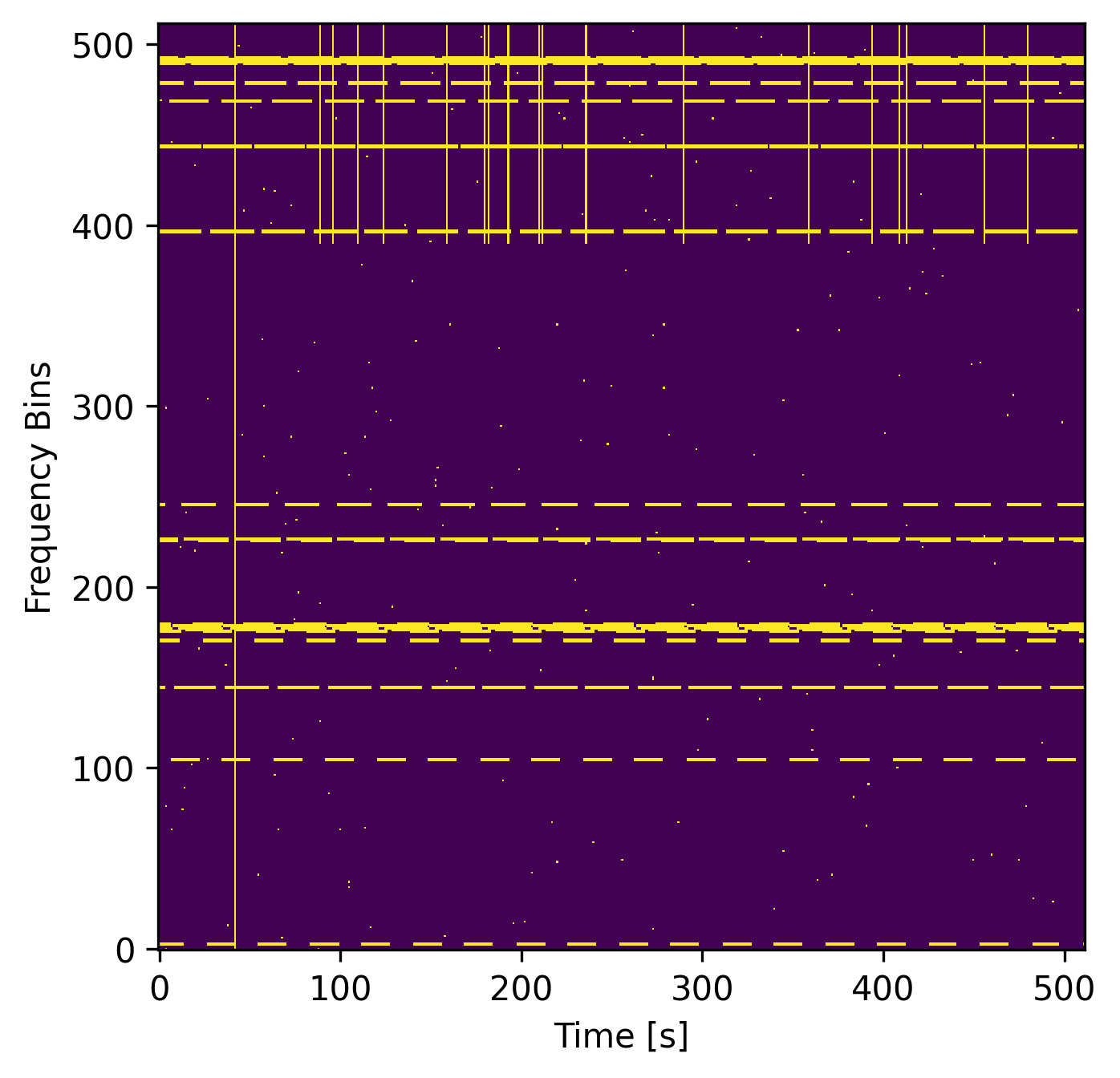}
        \caption{Ground-truth RFI mask.}
        \label{fig:res:mask}
    \end{subfigure}
    \caption{Example original $512 \times 512$ sized spectrogram, output of a transformer-readout direct-encoded LSM and the supervised ground-truth. It is clear that the LSM struggles with wide-band, short-duration RFI and periodic RFI.}
\label{fig:example}
\end{figure}
\section{Conclusions}
In this work, we demonstrate that liquid state machines can perform tasks beyond classification by tasking an LSM to detect RFI in a well-known synthetic dataset.
We tested multiple encoding techniques and increasingly complex readout layers, finding that direct exposure with rate-decoding and a transformer decoder readout layer provide the best performance. We also find that utilizing the liquid's spiking output as the memory for the transformer layer significantly improves performance, providing a hybrid of both spiking and artificial neural networks to handle a task previously unseen for LSMs.
While results lag the state of the art for trained SNNs and ANN2SNN conversions, this work recovers rate encoding as a viable scheme.
The relative computational simplicity of training LSMs continues to make them an appealing direction for investigation.
Moreover, this work is the first to employ second-order integrate and fire LiF neurons.
This work demonstrates SNNs' fundamental flexibility and introduces a biologically inspired network model for RFI-detection.
Future directions could expand the number of columns present in the reservoir to handle both slow and fast-varying RFI, test an SNN-based decoding layer using the membrane potential as a decoding method in addition to the encoding demonstrated here, and finally expand the learning paradigms to include unsupervised or partial evolution of the reservoir itself.
There is a growing understanding that SNNs can bring the computational power of connectionist methods while saving on energy usage, and this work adds to the body of work in the fledgling field of applying SNNs to astronomical sciences.
\FloatBarrier
\bibliographystyle{IEEEtran}
\bibliography{IEEEabrv,IJCNN-25} %
\end{document}